\documentclass[letterpaper, 10 pt, conference]{ieeeconf}
\usepackage[T1]{fontenc}
\usepackage{CJKutf8}
\usepackage{amsfonts}
\usepackage{amsmath}
\usepackage{siunitx}
\usepackage{booktabs}
\usepackage{multirow}
\usepackage{graphicx}
\usepackage{cite}
\usepackage{balance}

\usepackage{hyphenat}

\usepackage[dvipsnames]{xcolor}
\usepackage{pifont}
\IEEEoverridecommandlockouts                              

\title{\LARGE \bf
TGRPO: Fine-tuning Vision-Language-Action Model via Trajectory-wise Group Relative Policy Optimization
}

\author{Zengjue Chen$^{1}$, Runliang Niu$^{1}$, He Kong$^{1}$,
Qi Wang$^{1*}$, Qianli Xing$^{1}$, Zipei Fan$^{1}$%
\thanks{$^{1}$ School of Artificial Intelligence, Jilin University, Changchun, China.}%
\thanks{*Correspondence: \texttt{qiwang@jlu.edu.cn}}%
}

\begin{document}

\maketitle
\thispagestyle{empty}
\pagestyle{empty}

\begin{abstract}
Visual-Language-Action (VLA) models have demonstrated strong cross-scenario generalization capabilities in various robotic tasks through large-scale pre-training and task-specific fine-tuning. However, their training paradigm mainly relies on manually collected successful demonstrations, making it difficult to adapt to complex environments when encountering out-of-distribution (OOD) scenarios or execution biases. While Reinforcement Learning (RL) provides a closed-loop optimization framework via active trial-and-error mechanism, it suffers from sparse rewards, high variance, and unstable optimization in long-horizon robotic tasks. To address these limitations, we propose Trajectory-based Group Relative Policy Optimization (TGRPO), an online RL-based training framework for VLA models. TGRPO leverages task analysis generated by a large language model to automatically construct dense reward functions, providing fine-grained feedback to accelerate convergence and improve credit assignment. The core of our method is a group-based strategy that samples and normalizes multiple trajectories in parallel, reducing variance through relative comparison. By integrating trajectory-level and step-level advantage estimation, TGRPO captures both global and local optimization signals without relying on a value network. Experiments on four task categories of the LIBERO benchmark demonstrate that TGRPO achieves an average success rate of 80.7\%, which is 4.2\% higher than that of Supervised Fine-Tuning (SFT) and outperforms other representative RL-based post-training methods. 
\end{abstract}

\section{INTRODUCTION}
Vision-Language-Action (VLA)~\cite{black2024pi_0,rt1,kim2024openvla,liu2024rdt, Driess2023PaLMEAE,brohan2023rt,bu2025univla,qu2025spatialvla, patratskiy2025spatial, li2024cogact, shukor2025smolvla, kim2025fine, wang2025bitvla,  shi2025memoryvla, lin2025evo, liang2025discrete, lv2025f1, zhang2025align,huang2025graphcot} models have gradually emerged as a crucial approach to achieve general robotic policies, exhibiting robust performance in robotic tasks across diverse scenario~\cite{khazatsky2024droid,o2024open,fang2023rh20t,bu2025agibot}. Despite their successes, VLAs are trained solely on human-provided successful demonstrations and lacks the ability to learn from failures. As a result, it shows limited flexibility in real-world interactions, often remaining at the "action memorization" level rather than aligning actions with environmental context \cite{chu2025sft}. This not only restricts its capacity for autonomous exploration and self-correction, but also limits its performance in out-of-distribution (OOD) tasks and scenarios.
\begin{figure}[t]
\centering
  \includegraphics[width=\linewidth,scale=1.00]{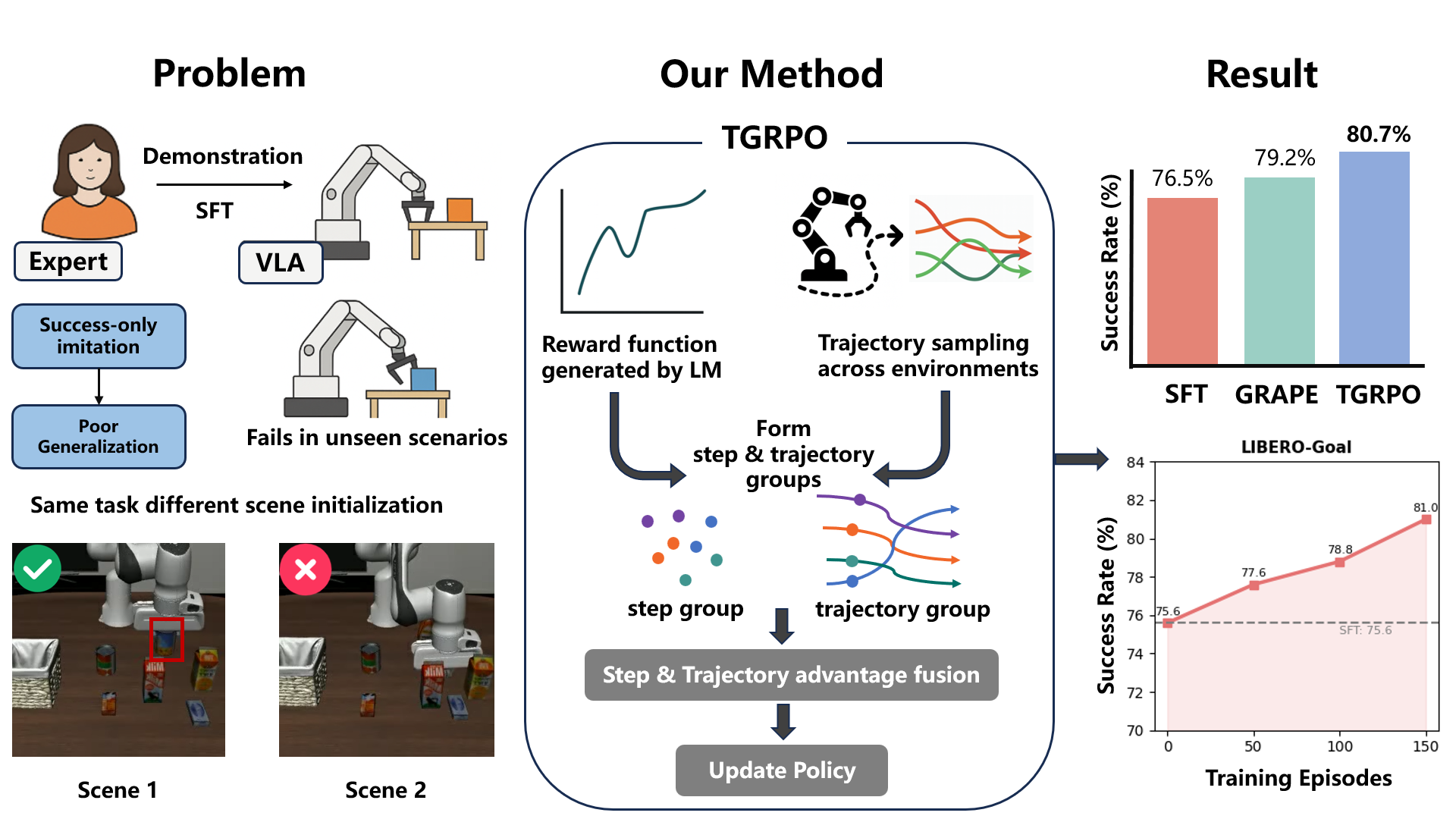}
  \caption{Problem: VLA models trained by SFT generalize poorly and fail in unseen scenarios. Method: TGRPO uses LLM-generated rewards and step–trajectory group fusion for stable policy learning. Result: On the LIBERO benchmark, TGRPO outperforms SFT and GRAPE.
  }
  \label{fig:Intro}
  \vspace{-0.1in}
\end{figure}

Reinforcement learning (RL) has been widely regarded as a natural and effective training paradigm that enables agents to optimize policies through interaction with the environment and to achieve autonomous behaviors in complex tasks~\cite{mehta2024waypoint,mcilvanna2022reinforcement,zhang2024gamma,huang2024mentor,luo2024precise}. A large body of work has successfully applied policy gradient methods, such as Proximal Policy Optimization (PPO)~\cite{schulman2017proximal} and Soft Actor-Critic (SAC)~\cite{haarnoja2018soft}, as well as value-based approaches to tackle robotic tasks, including motion control, grasping, and manipulation, demonstrating impressive performance in both simulated and real-world environments. However, PPO requires simultaneously training a value function network (Critic) to estimate state values and balances bias and variance through Generalized Advantage Estimation (GAE)~\cite{schulman2015high}, which can hinder training efficiency and effectiveness. At the same time, reward signals in real-world robotic tasks are often highly sparse, frequently reduced to binary success/failure feedback at the episode level, making exploration and optimization challenging. 

While researchers commonly adopt reward shaping or hand-crafted dense rewards to address this issue, such methods demand significant domain expertise and risk introducing reward bias, potentially diverting learning from the true task objective. Even with dense rewards, long-horizon robotic tasks pose further challenges: they involve multiple stages with diverse dynamics, where inconsistent reward scaling across subgoals can amplify gradient variance, slow convergence, destabilize training, and trap policies in suboptimal local minima. Agents may even overfit intermediate sub-goals with disproportionately large rewards while neglecting overall task success. Thus, effectively balancing reward distributions across stages and reducing training variance remain key bottlenecks for scaling RL in robotic learning.

In this paper, we propose Trajectory-wise Group Relative Policy Optimization (TGRPO), a critic-free training algorithm for Vision-Language-Action (VLA) models. Specifically, TGRPO leverages a large language model (LLM) to extract task-relevant object information from the environment and combines it with end-effector pose data from demonstrations to construct a dense, multi-stage reward function. Using this reward, the VLA model controls robots in parallel environments to sample diverse trajectories. These trajectories are then organized into groups at both the trajectory level and the step level: entire trajectories are grouped to capture global performance, while steps at the same timestep across trajectories are grouped to capture local consistency. Relative advantages are computed within each group and fused across levels, integrating local and global optimization signals for stable policy updates. The overall problem setting, proposed method, and experimental results are illustrated in Fig. 1. Experimental results show that TGRPO achieves superior success rate against both SFT fine-tuning and RL-based approaches for VLA models, demonstrating its effectiveness and generalization capability.

The main contributions are summarized as follows: (1) We introduce an online RL-based training framework for VLA models, addressing the fundamental limitation of training only from successful demonstrations and enabling VLA agents to actively explore and learn from interactions with the environment. This suits adaptive policy learning in dynamic robotic task scenario; (2) We design the TGRPO algorithm, a novel group-based policy optimization method integrating trajectory-level and step-level advantage estimation. Tailored to robot long-horizon multi-stage tasks, this integration fuses global and local optimization signals, avoiding single-level advantage estimation’s flaw in balancing global-task and local-action optimization; (3) We highlight the synergy of dense reward design and group-based optimization for robotic RL. Leveraging LLMs to parse tasks and generate stage-wise dense rewards, combined with the group-based strategy, TGRPO substantially improve the effectiveness of RL training for VLA models in complex robotic tasks.

\begin{figure*}[t]
\centering
  \includegraphics[width=\linewidth,scale=1.00]{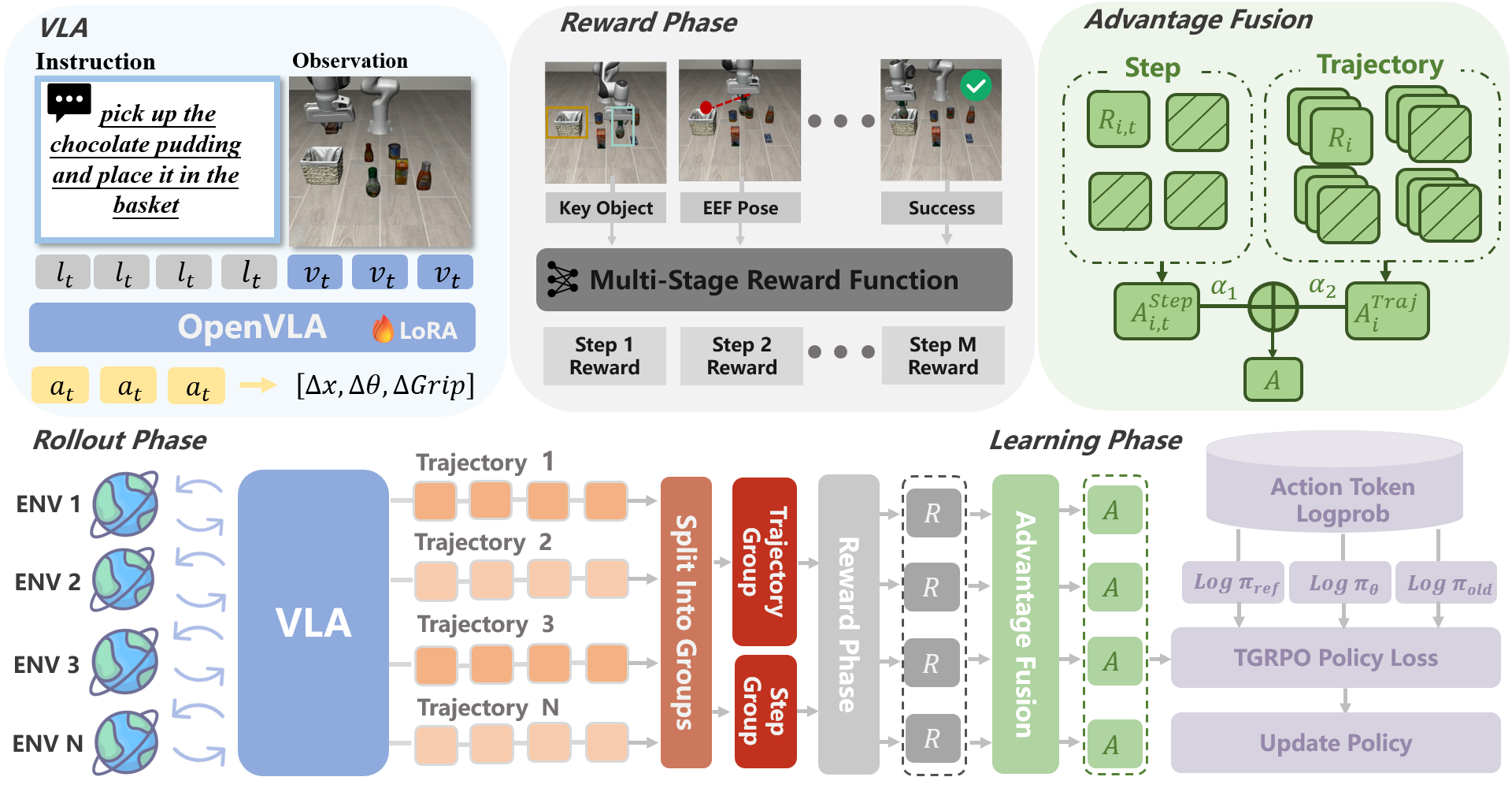}
  \caption{\textbf{Overview of the proposed Trajectory-wise Group Relative Policy Optimization.} Given natural language instructions and multimodal observations, the OpenVLA model produces action tokens to control the robot. Trajectories sampled in parallel across environments are evaluated with a multi-stage reward function, grouped for step-level and trajectory-level advantage estimation, and fused to update the policy.}
  \label{fig:Network}
\end{figure*}

\section{RELATED WORK}
\textbf{Vision-Language-Action Models. }
Existing VLA models can be mainly categorized into hierarchical robotic models and end-to-end robotic models. Hierarchical models, such as Rekep~\cite{huang2024rekep}, VoxPoser~\cite{huang2023voxposer}, Omnimanip~\cite{pan2025omnimanip} and Hi Robot~\cite{shi2025hi}, advocate dividing robotic models into high-level policies and low-level policies. The high-level policy handles understanding and reasoning about complex tasks, decomposing them into executable instructions for low-level policies, which then translate these instructions into corresponding robotic actions. In contrast, VLA model more commonly refers to end-to-end robotic models. Classic end-to-end VLA models like PaLM-E~\cite{Driess2023PaLMEAE}, RT-2~\cite{brohan2023rt}, and OpenVLA~\cite{kim2024openvla} typically use autoregressive large language models as their backbone, outputting discrete action tokens. OpenVLA employs SigLip~\cite{zhai2023sigmoid} and DINOv2~\cite{oquab2023dinov2} as its vision encoders and Llama2-7B~\cite{touvron2023llama} as its backbone. Through large-scale pre-training on the OXE open-source robotic dataset~\cite{o2024open}, the model gains the ability to comprehend both the current scene images and task-specific language instructions, subsequently generating corresponding action tokens as outputs. Recently, diffusion policy~\cite{chi2023diffusion} is integrated into VLAs, such as \(\pi_0\)~\cite{black2024pi_0}, GO-1~\cite{bu2025agibot}, and RDT-1B~\cite{liu2024rdt}. The \(\pi_0\) architecture builds on a pre-trained vision-language model (VLM) backbone (e.g., PaliGemma~\cite{beyer2024paligemma}) augmented with an "action expert" that uses flow matching to generate continuous action distributions for dexterous robotic control. Despite these  promising performance, VLA models exhibit limitations when applied to out-of-domain tasks. 


\textbf{Reinforcement Learning for VLA Fine-Tuning.} Due to the limitations of the aforementioned works, many studies have gradually turned their attention to using reinforcement learning to fine-tune VLA. A recent work proposed a VLA model fine-tuning method called ConRFT~\cite{chen2025conrft}, which combines offline behavior cloning, Q-learning, and online consistency policy training, and introduces human intervention to improve exploration efficiency and safety. Another study introduced the iRe-VLA framework~\cite{guo2025improving}, which iterates between reinforcement learning and supervised learning to enhance the performance of the VLA model. In addition, GRAPE~\cite{zhang2024grape} was proposed, which implicitly models rewards for successful and failed trials at the trajectory level to improve the generalization ability of the VLA model for new tasks. At the same time, it splits complex operation tasks into several stages and uses key points generated by large vision-language models and customized spatio-temporal constraints to automatically guide preference modeling. Different from these works, our approach does not require excessive human intervention, nor does it need to accumulate a large amount of offline datasets. It focuses more on online training with the robot, and the training time is significantly reduced. 

\section{PROBLEM FORMULATION}
In this paper, the policy optimization of VLA model $\pi_\theta$ is achieved by the online reinforcement learning. Unlike supervised fine-tuning (SFT), which relies solely on offline successful demonstrations, our work focuses on optimizing policies through direct interaction with the environment, thereby equipping VLA with the ability to learn from failures and self-correct. Formally, the task is modeled as a Markov decision process, defined by the tuple $(\mathcal{S},\mathcal{A},\mathcal{P},\mathcal{R},\mathcal{V},\mathcal{L})$, where $\mathcal{S}$ is the state space and $\mathcal{A}$ is the action space. $\mathcal{P}$ is a transition function. $\mathcal{R}$ is the reward function. $\mathcal{V}$ is the observation space captured from a third-person camera. $\mathcal{L}$ is a set of natural language instruction. In each timestep, the VLA model takes action $a \in A$ based on the observation $v \in V$ and corresponding $l\in L$.  
Then, the VLA model will receive  the reward $r$. In this paper, the reward is dense reward provided by an extra LLM, in which the reward function produces the reward at each timestep $t$ based on the positions of key objects provided by the simulation environment and the robot’s end-effector pose, i.e., the robot’s state $s_t$.  The overall objective of our policy is to maximize the cumulative trajectory reward $R = \sum_{t=1}^{M} r_t$, where $M$ denotes the shortest length among multiple trajectories.

\section{METHODOLOGY}


To better capture the characteristic of long-horizon robotic tasks, we propose Trajectory-wise Group Relative Policy Optimization (TGRPO), an online
reinforcement learning framework for post-training VLA fine-tuning. In concrete, TGRPO performs trajectory sampling across multiple environments initialized with identical states. During sampling, a multi-stage reward function designed by a large language model provides step-wise dense rewards. The collected trajectories are then organized into step-level and trajectory-level groups, where relative advantages are computed and fused to combine local and global optimization signals. The final loss is used to update the VLA policy effectively.
Fig. 2 shows the framework of our TGRPO.
\subsection{Multi-Stage Reward Design}
In our TGRPO framework, step-level advantages are computed before being fused with trajectory-level estimates. Simply propagating a terminal binary reward to all steps neglects the heterogeneous contributions of different actions. For example, in a failed trajectory, early steps may still achieve several sub-goals. Thus, we introduce a multi-stage reward function that provides fine-grained signals 

We leverage Claude 3.7 Sonnet to automatically decompose tasks into sub-stages from natural-language descriptions. For example, in the LIBERO-Object task of placing a tomato sauce bottle into a basket, the stages include approaching, grasping, moving, and placing. The LLM defines stage-wise rewards using spatial information of key objects and identifies the current stage at runtime. To further enhance efficiency, we add a shaping signal based on the end-effector pose: pre-collected successful poses are used as references, and the Euclidean distance to them provides additional guidance, encouraging alignment with expert demonstrations and accelerating task completion.

We first define the positions of the key objects in the task as $P_{object} \in \mathbb{R}^3$, where the subscript $object$ denotes the name of the task-relevant item. For example, in the task of placing a tomato sauce bottle into a basket, we have both $P_{tomato}$ and $P_{basket}$. Next, we define the set of reference key poses obtained from successful demonstrations as $\{P^1_{pose}, P^2_{pose}, \dots,P^j_{pose}\}$, $P^k_{pose} \in \mathbb{R}^3$. And each $P^k_{pose}=(x,y,z)$ represents the 3D Cartesian position of the robot end-effector at a critical step of the task, and $j$ denotes the total number of key poses. Based on the task description, together with the set of reference poses 
$P^k_{pose}$ and the task-relevant object positions $P_{object}$, the LLM automatically generates a multi-stage reward function. Therefore, the reward for each step in the trajectory should be defined as 
\begin{gather}
R_t=f_1(P_{object}(t),P^k_{pose})+f_2(P^k_{pose},s_t) .
\end{gather}
Here, $f_1(\cdot)$ and $f_2(\cdot)$, respectively denote the reward function based on key object information and the reward function based on key successful demonstration poses.
\begin{table*}[t]
\begin{center}
\setlength\tabcolsep{4pt}
\caption{Results on LIBERO (Mean success rate \%, Rank). Best results are indicated in bold.}
\label{tab:TGRPO and baselines}
\begin{tabular}{l cc cc cc cc cc}
\toprule
& \multicolumn{2}{c}{LIBERO-Spatial} 
& \multicolumn{2}{c}{LIBERO-Object} 
& \multicolumn{2}{c}{LIBERO-Goal} 
& \multicolumn{2}{c}{LIBERO-Long} 
& \multicolumn{2}{c}{Average} \\

\cmidrule(lr){2-3} \cmidrule(lr){4-5} \cmidrule(lr){6-7} \cmidrule(lr){8-9} \cmidrule(lr){10-11}
Method  & Success Rate & Rank & Success Rate & Rank & Success Rate & Rank & Success Rate & Rank & Success Rate & Rank \\
\midrule

Octo &
77.6 & 5 & 84.9 & 5 & 82.9 & 2 & 50.3 & 5 & 73.9   & 4.25\\

OpenVLA-SFT &
84.7 & 3  &  88.4 & 4 & 79.2 & 5 & 51.1 & 4 & 76.5 & 4\\

OpenVLA-DPO & 84.2 & 4 & 88.6 & 3 & 79.5 & 4 & 52.6 & 3 & 76.2 & 3.5  \\

GRAPE  & 88.5 & 2 & 92.1 & 2 &
\textbf{83.1} & \textbf{1} & 57.2 & 2 & 80.2 & 1.75 \\
\textbf{TGRPO (Ours)} &
\textbf{90.4}  & \textbf{1} & \textbf{92.2} &\textbf{1} &
81 & 3  & \textbf{59.2} & \textbf{1} & \textbf{80.7} & \textbf{1.5} \\

\bottomrule
\end{tabular}
\end{center}
\end{table*}
\begin{figure*}[t]
\centering
  \includegraphics[width=0.9\linewidth,scale=1.00]{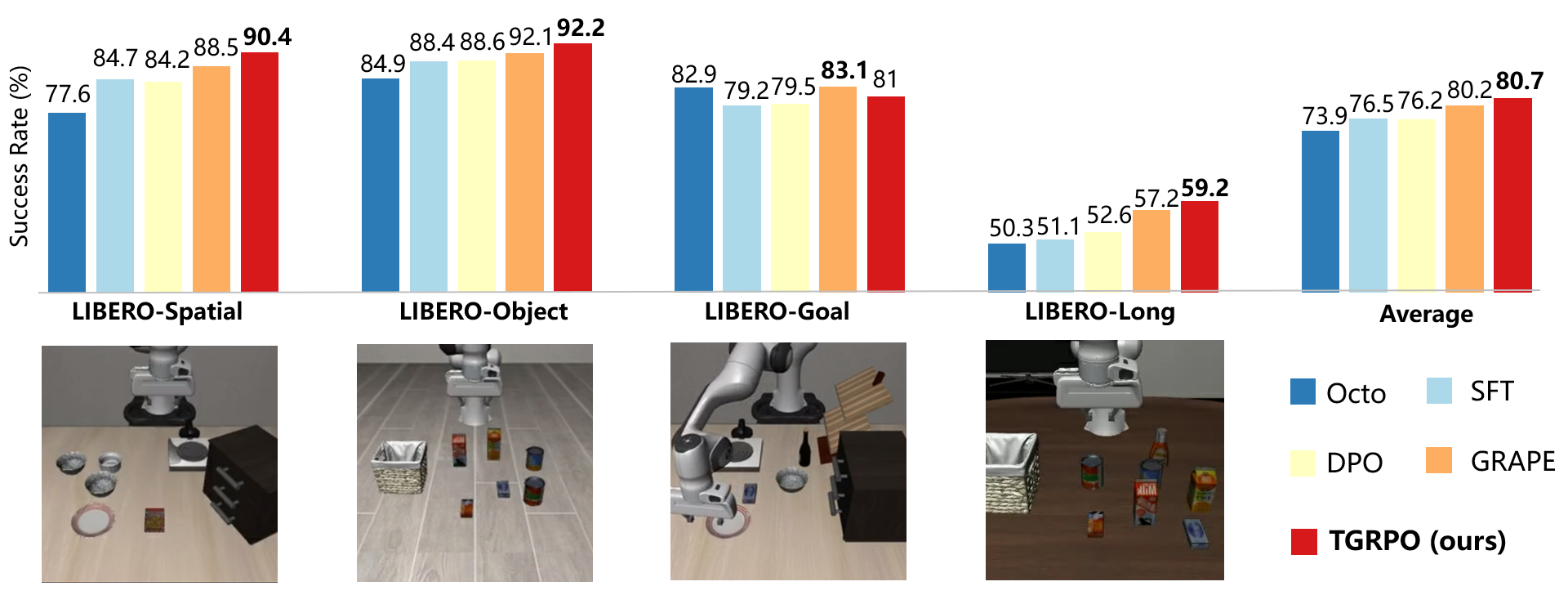}
  \caption{\textbf{Success rates of different methods on the LIBERO benchmark suites.} Our proposed TGRPO consistently outperforms baseline methods (Octo, SFT, DPO, GRAPE) across all task categories (Spatial, Object, Goal, Long), achieving the highest average success rate of 80.7\%.}
  \label{fig:Network}
\end{figure*}

\subsection{Trajectory-wise Group Relative Policy Optimization}
With the multi-stage reward function, we are able to assign meaningful reward signals to each step within a trajectory, which facilitates subsequent policy optimization. Our optimization approach extends the group-based strategy by refining both the grouping procedure and the computation of advantages in the final loss. Specifically, we incorporate trajectory-level and step-level advantage estimation to better capture the structure of long-horizon robotic tasks, thereby improving credit assignment and enhancing policy stability in sequential decision-making. Suppose we have $N$ trajctories, each of length $M$. The reward at step $t$ of trajectory $i$ is denoted as $R_{i,t}$. At this point, the cumulative sum of step rewards within each trajectory represents its total reward: $
R_i = \sum_{t=1}^{M} R_{i,t} $. We group together the steps at the same timestep $t$ from all $N$ trajectories, which allows us to compute the step-level relative advantage as:
\begin{gather}
    {A}_{i,t} = \frac{{R_{i,t}}-\frac{1}{N}\sum_{i=1}^{N} R_{i,t}}{\sqrt{\frac{1}{N-1} \sum_{i=1}^{N} (R_{i,t} - \frac{1}{N}\sum_{i=1}^{N} R_{i,t})^2})}.
\end{gather}
Next, we treat the entire set of $N$ trajectories as a single group to compute the relative advantage of each trajectory within the group. Formally, the trajectory-level relative advantage for trajectory $i$ is defined as:
\begin{gather}
    {A}_{i} = \frac{{R_i}-\frac{1}{N}\sum_{i=1}^{N} R_i}{\sqrt{\frac{1}{N-1} \sum_{i=1}^{N} (R_i - \frac{1}{N}\sum_{i=1}^{N} R_i)^2})}.
\end{gather}
Combining the two and weighting them with hyperparameters $ \alpha_1$ and $\alpha_2$, we obtain the final relative advantage for each step within every trajectory. Thus, the final relative advantage for step $t$ in trajectory $i$ is given as follows:
\begin{gather}
    {Adv}_{i,t} = \alpha_1 {A}_{i,t} +  \alpha_2 {A}_{i}  ,
\end{gather}
In our experiments, we empirically set $\alpha_1=0.3$ and $\alpha_2 = 0.7$ based on hyperparameter tuning.
Finally, we incorporate this relative advantage into a policy optimization objective. We define the importance sampling ratio between the updated policy $\pi_\theta$ and the old policy $\pi_{\theta_{\text{old}}}$ as:
\begin{gather}
\rho_{i,t} = \frac{\pi_{\theta}(a_{i,t} \mid s_{i,t})}{\pi_{\theta_{\text{old}}}(a_{i,t} \mid s_{i,t})}.
\end{gather}
The optimization objective follows a clipped surrogate formulation similar to GRPO, incorporating our dual-level relative advantage:
\begin{gather}
\begin{aligned}
\mathcal{J}_{\text{TGRPO}}(\theta) 
&= \mathbb{E}\!\left[q \sim P(Q), \{o_i\}_{i=1}^N \sim \pi_{\theta_{\mathrm{old}}}(O \mid q)\right] \\
&\quad \cdot \frac{1}{N} \sum_{i=1}^{N} \frac{1}{|o_i|} \sum_{t=1}^{|o_i|}
\Biggl\{
  \min \Biggl[
  \rho_{i,t} {Adv}_{i,t},\\
&\quad\quad 
    \operatorname{clip}\!\left(
\rho_{i,t},
      1-\epsilon,\,1+\epsilon
    \right) {Adv}_{i,t}
  \Biggr] \\
&\quad\quad\quad\quad 
  - \beta D_{\mathrm{KL}}\!\bigl[\pi_\theta \parallel \pi_{\text{ref}}\bigr]
\Biggr\}.
\end{aligned}
\end{gather}
Here, $o$ denotes a sampled trajectory, while $q$ represents the current task environment, which directly encompasses the sequence of visual observations and the task instruction as input. For the KL divergence term, we adopt an unbiased estimator:
\begin{gather}
\begin{aligned}
D_{\mathrm{KL}}\!\left[\pi_\theta \parallel \pi_{\text{ref}}\right] 
&= \frac{\pi_{\text{ref}}(a_{i,t} \mid s_{i,t})}{\pi_{\theta}(a_{i,t} \mid s_{i,t})}\\
&- \log \frac{\pi_{\text{ref}}(a_{i,t} \mid s_{i,t})}{\pi_{\theta}(a_{i,t} \mid s_{i,t})} - 1 .
\end{aligned}
\end{gather}

In summary, the TGRPO algorithm optimizes the policy by maximizing Eq. (6), thereby driving the continual improvement of the VLA model’s capabilities.

\subsection{VLA Post-training Framework in Simulation}
After improving the computation of relative advantages and deriving the corresponding optimization objective, we integrated these components into a complete online reinforcement learning framework for VLA post-training in simulation.

First, our overall framework trains a VLA model for a single task using reinforcement learning across multiple environments initialized with identical states. In this setup, the VLA executes the same task in parallel environments, sampling actions step by step until either one environment completes the task or all environments reach the maximum number of steps. During sampling, we incorporate the multi-stage reward function designed by the LLM described earlier, where each environment’s observations provide the necessary object positions and robot state information required for reward computation. Once the trajectories terminate simultaneously, they all share the same length, which facilitates consistent grouping for subsequent processing.

After collecting multiple trajectories, they are organized into a trajectory-level group, where relative advantages are computed within the group according to Eq. (3), yielding trajectory-level relative advantages. Similarly, since all trajectories terminate at the same timestep (ensuring that every step can be grouped), we extract step-level data across trajectories (e.g., rewards and log probabilities of actions), and group together steps at the same timestep to form step-level groups. Step-level relative advantages are then computed using Eq. (2).

Finally, the trajectory-level and step-level advantages are combined with weights as defined in Eq. (4), and the resulting objective is optimized using the loss function in Eq. (6) to update the model parameters.

\section{EXPERIMENTS}
\begin{table*}[t]
\centering
\caption{Ablation studies on LIBERO-Goal (Success rate \%). Best results are indicated in bold.}
\label{tab:ablation_study}
\begin{tabular}{l ccccccccccc}
\toprule
& Task0 & Task1 & Task2 & Task3 & Task4 & Task5 & Task6 & Task7 & Task8 & Task9 & Avg. \\
\cmidrule(lr){2-12}
       Method & SR &  SR & SR & SR & SR & SR & SR & SR & SR & SR & SR\\
\midrule
SFT & 86 & 76 & 90 & 74 & 92 & 92 & 98 & 92 & 92 & 92 & 88.4 \\
TGRPO w/o Trajectory-level Adv. & 88 & 56 & 86 & 60 & 92 & 82 & 92 & 92 & 92 & 60 & 80.2 \\
TGRPO w/o Step-level Adv. & 78 & 78 & 98 & 58 & 94 & 82 & 96 & 96 & 92 & 96 & 86.8 \\
\textbf{TGRPO (Ours)} & \textbf{88} & \textbf{82} & \textbf{98} & \textbf{76} & \textbf{98} & \textbf{94} & \textbf{98} & \textbf{98} & \textbf{94} & \textbf{96} & \textbf{92.2} \\
\bottomrule
\end{tabular}
\end{table*}
\subsection{Tasks and Experimental Setup}
In our experiments, we adopt OpenVLA~\cite{kim2024openvla} as the base VLA model and fine-tune it with LoRA during reinforcement learning training. All reinforcement learning updates are performed using the AdamW optimizer with a fixed learning rate of $1 \times 10^{-5}$. Consistent with the experimental setup in ~\cite{zhang2024grape}, we employ the LIBERO~\cite{liu2023libero} robotics simulator as our testbed. LIBERO provides a clean, Python-style API that grants direct access to object states, robot joint angles, and other scene information, enabling us to conveniently construct dense and informative reward functions. LIBERO is a benchmark for lifelong robot learning comprising four task suites: spatial, object, goal, and long, targeting generalization across spatial locations, object categories, task goals, and long-horizon complexity, respectively. Each suite contains 10 tasks. During training and evaluation, we sample one task at a time, run it in four parallel environments, and evaluate performance over 50 test episodes with average success rates reported per task and per suite. All experiments are conducted on a single NVIDIA A100 GPU.


\begin{figure}[tb]
\centering
  \includegraphics[width=0.95\linewidth,scale=1.00]{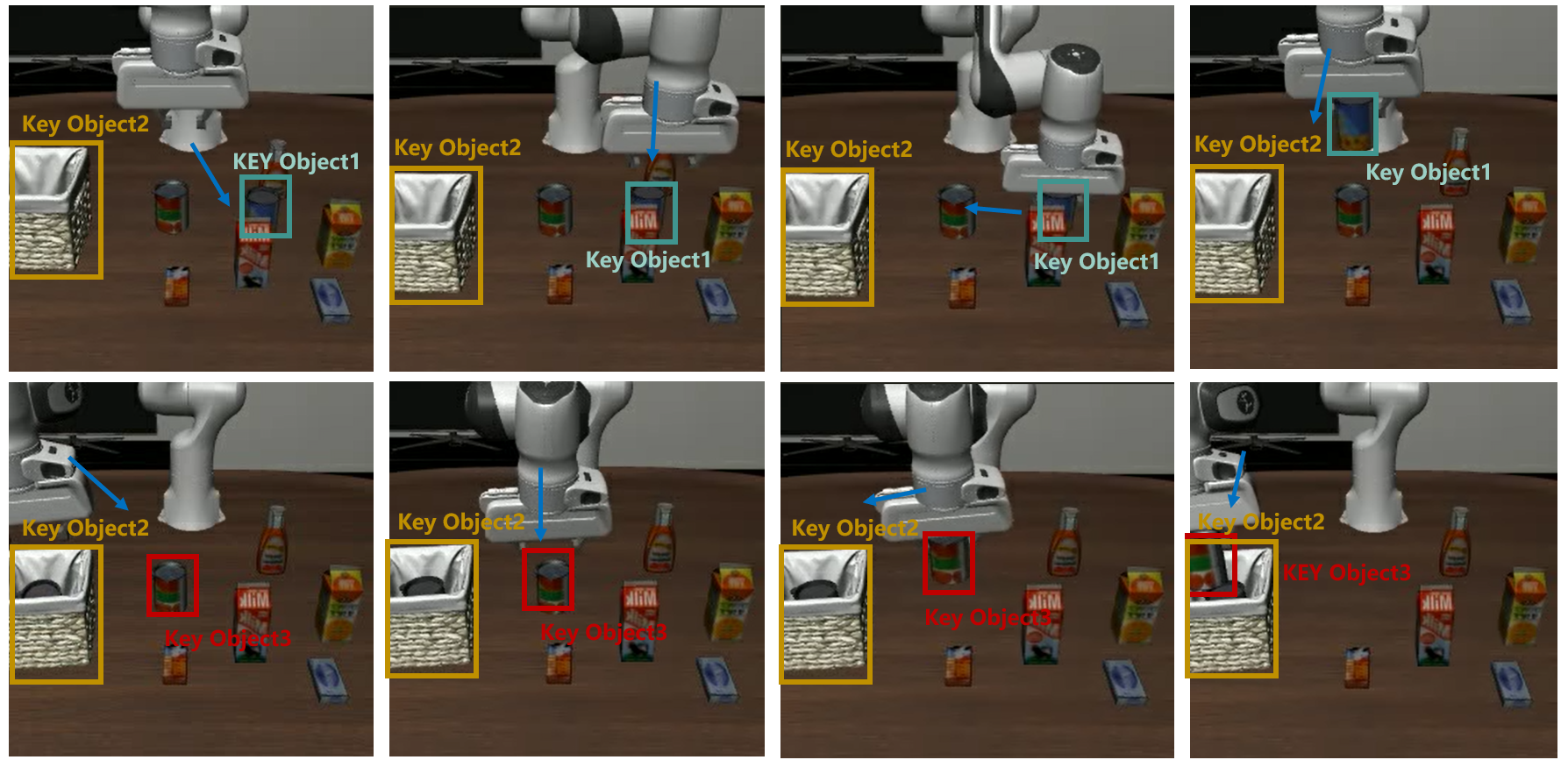}
  \caption{\textbf{An example from LIBERO-Long.} A successful trajectory trained with TGRPO. At each sampling step, rewards are assigned based on the states of key objects and the robot. The illustrated task is "put both the alphabet soup and the tomato sauce in the basket", where the key objects are alphabet soup, tomato sauce, and basket.}
  \label{fig:Gaussian}
\end{figure}

\subsection{TGRPO Evaluation on LIBERO}
\textbf{The performance of TGRPO on LIBERO.} TGRPO is a fine-tuning approach for VLA models. Accordingly, for baseline selection, we first include methods that also follow the fine-tuning paradigm, such as SFT, DPO, and GRAPE. In addition, to further highlight the effectiveness of our method, we incorporate several independent models that are not based on post-training as additional baselines. As shown in Table 1 and Fig. 3, our proposed TGRPO model achieves a significantly higher average success rate across the four LIBERO task suites compared to all baselines, demonstrating its superior performance in fine-tuning VLAs on new tasks. Specifically, TGRPO outperforms standard SFT fine-tuning by 4.2\% and also surpasses other reinforcement learning–based methods such as DPO and GRAPE. Notably, on the LIBERO-Long suite that involves long-horizon complex tasks, TGRPO achieves an average success rate that is 8.1\% higher than SFT. This result highlights that step-wise reward assignment and integration of trajectory- and step-level signals allow TGRPO to better enable VLA models to explore and learn effectively in challenging long-horizon scenarios.

\textbf{Ablation study.} Table 2 presents ablation results on the LIBERO-Object suite. Compared with the SFT baseline, our TGRPO framework achieves the highest success rate of 92.2\%, demonstrating the benefit of integrating both trajectory-level and step-level advantages. Removing trajectory-level advantages leads to a noticeable performance drop, highlighting their importance for capturing global task progress across trajectories. Similarly, removing step-level advantages results in an even larger decrease, indicating that fine-grained step-wise feedback is critical for effective credit assignment in long-horizon tasks. These results prove that the dual-level advantage design is essential for the stability and effectiveness of TGRPO.

\textbf{Case Study.} To further validate the effectiveness of TGRPO, we present a case study on the LIBERO-Long suite. The selected task, put both the alphabet soup and the tomato sauce in the basket, requires the robot to complete multiple sub-goals in sequence, making it a representative long-horizon challenge task. As shown in Fig. 4, the environment provides key object states (alphabet soup, tomato sauce, and basket) together with the robot’s state, which are used by our multi-stage reward function to assign step-wise feedback. This enables the policy to recognize partial progress such as approaching or grasping the correct object. By integrating both trajectory-level and step-level advantages, TGRPO effectively combines global task progress with fine-grained corrective signals.
\begin{figure}[tb]
\centering
  \includegraphics[width=\linewidth,scale=1.00]{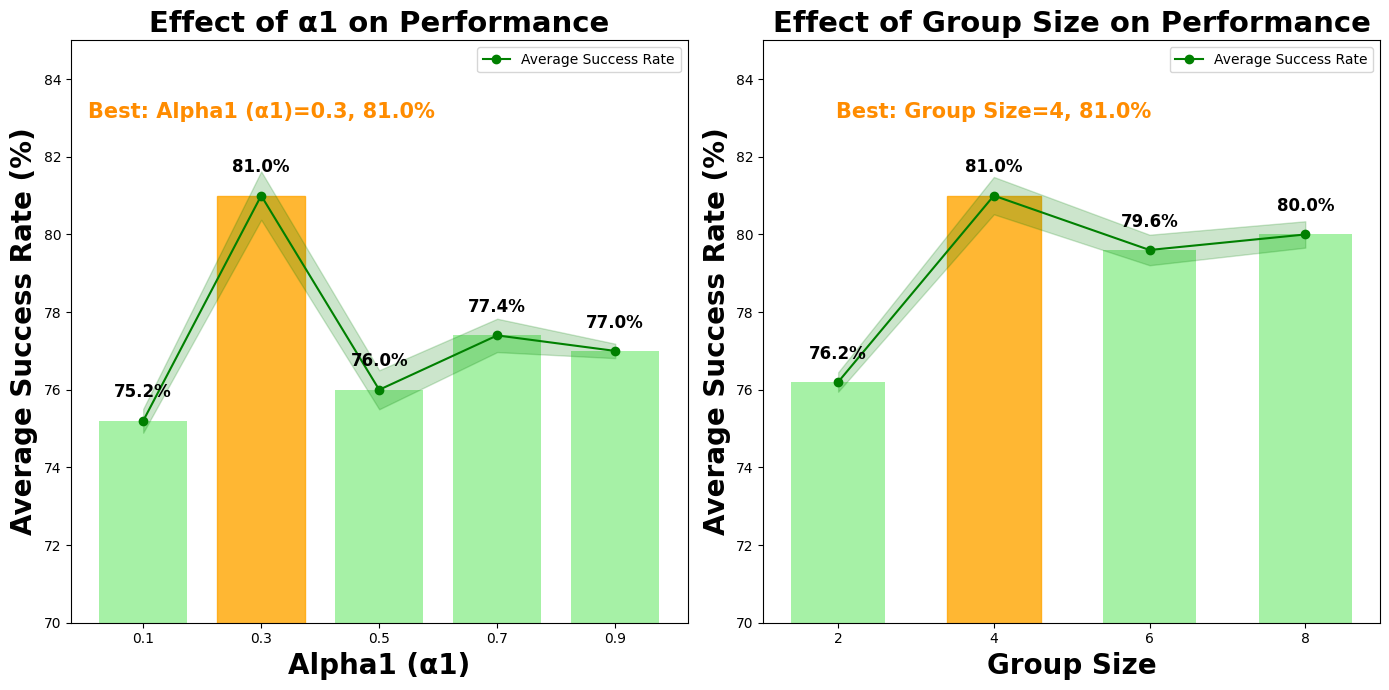}
  \caption{\textbf{Hyperparameter analysis of TGRPO on the LIBERO-Goal suite.} We evaluate the average success rate across different settings of the weighting coefficients $(\alpha_1,\alpha_2)$, which balance step-level and trajectory-level advantages. The best performance is achieved at $(0.3,0.7)$, yielding an average success rate of 81.0\%. }
  \label{fig:Gaussian}
\end{figure}
\subsection{Hyperparameter Analysis of Advantage Weights}
In TGRPO, the final relative advantage is defined as a weighted combination of step-level and trajectory-level advantages, parameterized by coefficients $\alpha_1$ and $\alpha_2$ with the constraint $\alpha_1 +\alpha_2 = 1$. To examine the impact of these weights, we conducted a set of hyperparameter analysis experiments on the LIBERO-Goal suite, which consists of long-horizon manipulation tasks. The results are showed in Fig. 5. We find that the performance of TGRPO is sensitive to the choice of $\alpha_1$. Specifically, setting $\alpha_1 =0.3$ and $\alpha_2 = 0.7$ yields the highest average success rate of 81.0\%, clearly outperforming other configurations. When $\alpha_1$ is too small (e.g.,0.1), the model underutilizes step-level advantages and fails to exploit fine-grained reward feedback, leading to weaker credit assignment. On the other hand, when $\alpha_1$ is too large, trajectory-level signals are overshadowed, reducing stability in long-horizon training. These results highlight that for long-horizon tasks in LIBERO-Goal, trajectory-level advantages serve as the dominant factor in stabilizing policy learning, while step-level advantages complement them by providing detailed guidance. A balanced configuration, $\alpha_1 =0.3$ and $\alpha_2 = 0.7$ , proves to be the most effective setting for maximizing the performance of TGRPO.

\subsection{Impact of Group Size on Policy Performance} 
The group size $N$ is the number of parallel trajectories sampled and aggregated per update in our framework. Larger groups may reduce variance and stabilize normalization at the cost of increased training time, whereas smaller groups can lead to noisier estimates and potentially degrade performance. To investigate the impact of group size in TGRPO, we evaluate $N \in \{2,4,6,8\}$ on the LIBERO-Goal suite. As shown in Fig. 5, setting group size too small $(N=2)$ leads to poor overall performance, with an average success rate of only 76.2\%. This suggests that insufficient trajectories per group fail to provide reliable relative comparisons, resulting in unstable optimization. When the group size increases to $N \geq 4$, the performance differences across configurations become relatively minor. Among these, $N=4$, achieves the best overall performance with an average success rate of 81.0\%, slightly outperforming larger group sizes while requiring fewer trajectories. This finding indicates that a moderate group size $N=4$ strikes the best balance between accuracy and efficiency, achieving stable performance while reducing training cost and improving the overall efficiency of VLA post-training.

\section{CONCLUSION}

In this work, we propose TGRPO, an online RL framework for fine-tuning Vision-Language-Action models that overcomes the limitations of supervised learning by enabling failure-aware policy improvement through environmental interaction. TGRPO uses LLM to automatically generate dense multi-stage rewards and introduces a group-based dual-level advantage estimation mechanism—combining trajectory-level and step-level signals—to enhance credit assignment and training stability without a value network. Experiments on LIBERO show TGRPO achieves an 80.7\% average success rate, outperforming SFT and prior RL methods. Ablations validate the necessity of both advantage levels, and hyperparameter studies confirm the optimal weighting for long-horizon tasks. TGRPO offers a scalable and efficient solution for adaptive VLA fine-tuning, reducing reliance on human engineering. Future work will extend TGRPO to real-world and multi-task settings.










\bibliographystyle{IEEEtran}
\balance
\bibliography{root}
\end{document}